\documentclass[11pt]{article}
\usepackage{acl2015}
\usepackage{times}
\usepackage{url}
\usepackage{latexsym}
\usepackage{booktabs}
\usepackage{xcolor}
\usepackage{soul}
\usepackage{float}
\usepackage{fourier} 
\usepackage{array}
\usepackage{makecell}

\title{Myers-Briggs Personality Classification and Personality-Specific Language Generation Using Pre-trained Language Models}

\author{Sedrick Scott Keh \\
    The Hong Kong University of \\
    Science and Technology \\
  {\tt sskeh@connect.ust.hk} \\\And
  I-Tsun Cheng \\
    The Hong Kong University of \\
    Science and Technology \\
  {\tt ichengaa@connect.ust.hk} \\}

\date{}

\begin{document}
\maketitle
\begin{abstract}
The Myers-Briggs Type Indicator (MBTI) is a popular personality metric that uses four dichotomies as indicators of personality traits. This paper examines the use of pre-trained language models to predict MBTI personality types based on scraped labeled texts. The proposed model reaches an accuracy of $0.47$ for correctly predicting all 4 types and $0.86$ for correctly predicting at least 2 types. Furthermore, we investigate the possible uses of a fine-tuned BERT model for personality-specific language generation. This is a task essential for both modern psychology and for intelligent empathetic systems.
\end{abstract}

\section{Introduction}
Proposed by psychoanalyst Carl Jung, the Myers-Briggs Type Indicator (MBTI) is one of the most commonly-used personality tests of the 21st century. Its prevalence ranges from casual Internet users to large corporations who use it as part of their recruitment process \cite{doi:10.1177/107179190301000106}. This test uses four metrics to capture abstract ideas related to one's personality. These four dichotomies are as follows:

\begin{itemize}
    \item introversion vs extroversion (I vs E)
    \item sensing vs intuition (S vs N)
    \item thinking vs feeling (T vs F)
    \item perception vs judging (P vs J)
\end{itemize}

The MBTI combines four letters to come up with an overall personality type. In total, there are $2^4=16$ personality types. Each of these types corresponds to a unique personality class. 

Overall, personality classification has many complexities because there are countless factors involved. Furthermore, even a human being may not be able to accurately classify a personality given a text. However, we hypothesize that using pre-trained language models might allow us to pick up on various subtleties in how different personality types use language. The objective of this paper is thus two-fold: given a set of labeled texts, first, we want to train a model that can predict the Myers-Briggs type, and second, use this model to generate a stream of text given a certain type.

\section{Related Work}
\subsection{Personality Classification Systems}
The earliest forms of personality recognition systems employed a combination of SVM and feature engineering \cite{rangel2016overview}. Other methods also utilized part-of-speech frequencies \cite{litvinova2015using}. More recently, newer systems have begun to use deep learning techniques such as convolutional neural networks and recurrent neural networks \cite{7960065}. These have the advantage of being able to easily extract meaningful features, as well as capture temporal relationships between neighboring context words.

Today, state-of-the-art systems for personality classification employ a wide combination of deep learning techniques. C2W2S4PT is a model that combines word bidirectional RNN and character bidirectional RNN to create hierarchical word and sentence representations for personality modeling \cite{liu-etal-2016-recurrent}. Other models also incorporate audio and video inputs in CNNs to output more accurate predictions \cite{kampman-etal-2018-investigating}.

\subsection{Pre-trained Language Models}
More recently, pre-trained language models such as BERT have begun to emerge  \cite{DBLP:journals/corr/abs-1810-04805}. BERT involves bidirectional training of transformers on two tasks simultaneously. These tasks are masked language modelling and next sentence prediction. BERT has led to numerous improvements in various areas of natural language processing, such as question answering and natural language inference \cite{zhangxu}.

However, to this day, very few research has been done on applying pre-trained language models on personality classification. As such, this paper introduces the use of BERT on classifying personality. Furthermore, we also extend this by training the models to be able to generate sentences given a type of personality. Such a feature will be helpful especially for empathetic dialogue systems.

\section{Data Scraping and Preprocessing}
\subsection{Scraping}
For MBTI personality types, there is no standard dataset. Previous studies on the topic mostly scraped datasets from various social media platforms, including Twitter \cite{inproceedings} and Reddit \cite{gjurkovic2018reddit}.

For this paper, we scraped posts from the forums in \url{https://www.personalitycafe.com}. PersonalityCafe's MBTI forums are divided into 16 distinct sections (one for each MBTI type). We also analyzed and discovered that over 95\% of users who post in a particular section identify as a member of that type, so these posts serve as a good general representation of how people of these personality types communicate. 

When scraping from this forum, we only considered posts that were over 50 characters long, since posts that are too short likely do not contain any meaningful information. Overall, we scraped the 5,000 most recent posts from each forum, resulting in a sample size of 68,733 posts and 3,827,558 words. (Some forums had less than 5,000 posts). This data was divided in an 85-15 train-test ratio.

\subsection{Cleaning and Preprocessing}
For data cleaning, we first removed all the symbols that were not letters, numbers, or punctuations, and then put spaces after punctuations to separate them as new tokens. We separated things like 're or 'll to be new tokens (ex. you're = you + 're). Lastly, we converted all tokens to lowercase.

Another important preprocessing step was to get rid of the instances where MBTI types were explicitly mentioned, and replace them with placeholder "$<$type$>$", as these may distort the task or make the task too easy for the classifier. 

\section{Classification Methodology}

\subsection{Tokenization}
For tokenization, we used BERT's custom tokenizer, which includes masking and padding the sentences. Sentences that were too long were truncated by the indicated maximum sequence length, and sentences that were too short were padded with zeros. For classification tasks such as this, the start of sentence was indicated with special token "[CLS]" and end of sentence with special token "[SEP]". This works because the model has been pre-trained and we are only fine-tuning over the pre-trained model \cite{DBLP:journals/corr/abs-1810-04805}.

\subsection{BERT Model Fine-Tuning and Training}
The fine-tuning model was created using Pytorch's custom "BertForSequenceClassification" model. Since the model was already pre-trained on 110 million parameters, the fine-tuning part basically involves training the final set of layers on our own corpus of inputs and outputs. The main purpose of fine-tuning is to allow the BERT bidirectional transformer model to adapt to our corpus and classification task, since the model itself was pre-trained on different tasks. Here, the main architecture for the BERT model's fine-tuning for the sequence classification task consists of 12 hidden layers with hidden size 768 and a dropout probability of 0.1, as well as incorporated attention functions with attention dropout of 0.1.

In the training part, we generated a dataloader and iterated over it. Training was done by batch (batch size 32). For the loss function, a cross entropy loss was used, and for the optimizer, we used the BertAdam optimizer with warmup proportion 0.1 and weight decay 0.01.

\section{Classification Results and Analysis}
\subsection{Hyperparameter Optimization}
There were 3 main parameters that we examined, namely learning rate, maximum sequence length, and number of epochs. Other parameters were mostly kept constant: we used bert-base-uncased model, training batch size of 32, evaluation batch size of 8, and warmup proportion of 0.1. The results are highlighted in the table in the following page. 

\begin{table}[h]
\centering
\begin{center}
\begin{tabular}{ |c|c|c|c| }
	\hline
	\textbf{Learn. Rate} & \textbf{Max Seq.} & \textbf{Epochs} & \textbf{Acc.} \\
	\hline
	0.001 & 128 & 5 & 0.0972 \\
	\hline
	0.0001 & 128 & 5 & 0.4701 \\
	\hline
	0.00001 & 128 & 5 & 0.4135 \\
	\hline
	0.000001 & 128 & 5 & 0.1739 \\
	\hline
	0.0000001 & 128 & 5 & 0.0901 \\
	\hline
	0.00001 & 128 & \textbf{30} & \textbf{0.4797} \\
	\hline
	0.00001 & \textbf{64} & 5 & 0.4138 \\
	\hline
	0.00001 & \textbf{256} & 5 & 0.4146 \\
	\hline
\end{tabular}
\end{center}
\caption{Classification accuracy of different parameter combinations.}
\label{tab:label}
\end{table}

\subsection{Discussion}

The best accuracy of the model was 0.4797, which was achieved with a learning rate of $10^{-5}$, maximum sequence length 128, and 30 epochs. This model vastly outperforms other models and previous benchmarks. Below is a table of how this BERT model performs compares to previous studies done on MBTI personality classification:

\begin {table}
\centering
\begin{center}
    \begin{tabular}{ |c|c|c|}
      \hline
      \thead{Method} & \thead{Dataset} & \thead{Acc.} \\
      \hline
      \makecell{Logistic Reg \\ (Plank \& Hovy, \\ 2015)} & \makecell{Twitter (2.1 \\ million tweets)} & 0.190 \\
      \hline
      \makecell{SVM \\ (Gjurkovic \& \\ Snajder, 2018)} & \makecell{Reddit \\ (22.9 million \\ comments)} & 0.370 \\
      \hline
      \makecell{LSTM \\ (Cui \& Qi, 2018)} & \makecell{Kaggle dataset \\ (8675 sentences)} & 0.380 \\
      \hline
      BERT & \makecell{PersonalityCafe \\ forums (68k posts)} & 0.479 \\
      \hline
    \end{tabular}
\end{center}
\caption{Classification accuracy of different parameter combinations.}
\label{tab:label}
\end{table}

Additionally, from the results of Table 1 (different parameter combinations), we can also analyze the effects of various parameters to the overall accuracy. First, we see that the learning rate plays a very significant role in the accuracy of the model. As we lower learning rate, the accuracy increases then begins to decrease. At learning rate levels of $10^{-3}$ and $10^{-7}$, accuracy was around 0.09, which was basically close to random guessing.

For the number of epochs, there is also a relatively significant increase in accuracy: when the number of epochs were increased from 5 to 30 for the $10^{-5}$ learning rate model, the accuracy increased from 0.4135 to 0.4797. Meanwhile, for the maximum sequence length, the difference is negligible, as changing from 128 to 64 or 256 only resulted in accuracy changes of less than 0.01.

\subsection{Further Results: Other Ways to Measure Accuracy}
Aside from the simple measure of accuracy (exact match of all 4 letters), since the MBTI personality consists of 4 letters, we can also consider other the number of correctly classified letters (personality categories) per prediction. This makes sense because for instance, if the true personality is INTJ, then a prediction of INTP (3 matches) would be a much better prediction as compared to a prediction of ESTP (1 match). Thus, for our personality classifier, we also consider the number of correctly classified letters as a measure of accuracy:

\begin{table}[h]
\centering
\begin{center}
\begin{tabular}{ |c|c|c|c| }
	\hline
    \begin{tabular}{@{}c@{}} \textbf{At least} \\ \textbf{1 match} \end{tabular} & 
    \begin{tabular}{@{}c@{}} \textbf{At least} \\\textbf{2 matches} \end{tabular} & 
    \begin{tabular}{@{}c@{}} \textbf{At least} \\ \textbf{3 matches} \end{tabular} & 
    \begin{tabular}{@{}c@{}} \textbf{At least} \\ \textbf{4 matches} \end{tabular} \\
	\hline
    0.9813 & 0.8573 & 0.6606 & 0.4797 \\
	\hline
\end{tabular}
\end{center}
\caption{Number of correctly predicted letters}
\label{tab:label}
\end{table}

From the table above, we can see that almost all the time, there is at least 1 match. Probabilistically, the expected number of matches is \textbf{2.9789}.

In addition, we can also consider the accuracy of the predictions each of the letter categories, to see if the BERT language model works well as a binary classifier:

\begin{table}[h]
\centering
\begin{center}
\begin{tabular}{ |c|c|c|c| }
	\hline
	\textbf{E/I} & \textbf{N/S} & \textbf{F/T} & \textbf{P/J} \\
	\hline
	0.7583 & 0.7441 & 0.7575 & 0.7190 \\
	\hline
\end{tabular}
\end{center}
\caption{Individual category accuracies}
\label{tab:label}
\end{table}

From the table above, we notice that "E/I" and "F/T" have the highest distinctions, which indicates that these two are the classes that the BERT model finds easiest to discern. Comparatively, the model has a relatively harder time differentiating between "P/J", although by not that large of a difference. These observations are consistent across other studies \cite{cuiqi} and generally reflect observations about the MBTI metric.

Overall, the performance as a binary classifier is actually not that high, as some other models such as SVM were able to achieve an accuracy of around 0.80 \cite{gjurkovic2018reddit}. We believe that with more training data and using a larger BERT model (bert-large), we can further increase the accuracy of our binary classifier model.

\section{Language Generation from Given Personality Type}
\subsection{Methodology}
In this task, we want to be able to generate a stream of text from a given MBTI personality type. The language generation is built using the Pytorch “BertForMaskedLanguageModelling” custom model, which consists of the BERT transformer with a fully pre-trained masked language model. Similar to the case in the sequence classification task, the main architecture of the pre-trained BERT model contains 12 layers, a hidden size of 768, and around 110 million parameters. However, while the BERT model used in classification task focuses only on lower case words, the BERT model here accounts for upper case words too, since we want the language generation model to train and learn when to generate upper-case and lower-case words.

The tokenizer implemented in this language model uses the same method as that in classification.

For each personality type, we train the language model on the corresponding texts we scraped. We compute the losses after each epoch and gather the final losses of the experiment run for each personality type as the results.

\subsection{Language Generation Results}

\begin{table}[H]
\centering
\begin{tabular}{|cc|cc|}
\hline
Type & Loss & Type & Loss \\ \hline
ENFJ & 0.01591 & INFJ & 0.032599 \\
ENFP & 0.021193 & INFP & 0.028531 \\
ENTJ & 0.02907 & INTJ & 0.028092 \\
ENTP & 0.030716 & INTP & 0.028124 \\
ESFJ & 0.017829 & ISFJ & 0.027062 \\
ESFP & 0.016334 & ISFP & 0.025123 \\
ESTJ & 0.016708 & ISTJ & 0.02662 \\
ESTP & 0.025886 & ISTP & 0.0239 \\ \hline
\end{tabular}
\caption{Language Generation Results}
\label{table}
\end{table}

For the hyperparameter settings, we used a training batch size of 16, learning rate of $3 \cdot 10^{-5}$, number of training epochs of 10, max sequence length of 128, and a warmup proportion of 0.1. We found this to be quite optimal in training the language model of different personality types as the loss decreases consistently. The results after 10 epochs for each personality type are shown above.

\subsection{Discussion}
The results show that ENFJ, ESFJ, ESFP, ESTJ all have the lowest loss among all the personality types as their losses are all under 0.02 after training for 10 epochs. All of these types contain “E” as their value of the first dichotomy. Interestingly, this means BERT is better at generating language for extroverted personalities as opposed to introverted personalities. This might be due to the fact that there are more data for extroverted personalities as they tend to be more active on forums and thus BERT is more capable of mimicking them.

Here is an example of generated sentence trained on ENFJ that has the lowest loss of all. The red sentence is input and the highlighted sentence is generated.

\begin{quote}
``\textcolor{red}{I have no idea if he feels the same way, and I am too afraid to press it.} \hl{Our relationship is very special to too lose them not much reasons if for anyone if or [UNK] or so Kahn else anything which is public or now not is or or  ...}''  
\end{quote}

Comparing between the pairs that differ only by the second dichotomy "N/S" (for example, ENFP and ESFP) indicates that S usually results in lower loss than its counterpart. Intuition personalities are generally more less dependent on the senses and logic, while sensing personalities usually rely on more facts, suggesting that BERT can more easily generate more sensical and logical language as opposed to abstract language.

For the last two dichotomies, "F/T" and "J/P", they do not impact the performance of the language model that much. The first two dichotomies are more dominant in affecting the results of language generation.

\section{Conclusion and Future Work}
In this paper, we explored the use of pre-trained language models (BERT) in personality classification. For the classification task, the proposed model achieved an accuracy of 0.479. Furthermore, for the language generation, we achieved losses of around 0.02. Possible improvements to this model would include using larger and cleaner datasets. If computation and memory resources permit, we can also use a larger bert model (bert-large trained on 340 million parameters) instead of bert-base.

\bibliography{main}
\bibliographystyle{acl}

\end{document}